\documentclass[a4paper]{article}
\usepackage{CNVSRC2023}
\usepackage{epsfig,amssymb,amsmath,graphicx}
\usepackage{booktabs}
\usepackage{multirow}
\usepackage{amssymb}
\usepackage{pifont}
\usepackage{hyperref}

\ninept

\title{The NPU-ASLP System Description for Visual Speech Recognition in CNVSRC 2024}

\name{He Wang, Lei Xie$^{\ast}$ \thanks{*Corresponding author}}

\address{Audio, Speech and Language Processing Group (ASLP@NPU), School of Computer Science,\\Northwestern Polytechnical University, Xian, China\\
{\small \tt hwang2001@mail.nwpu.edu.cn}}%

\providecommand{\keywords}[1]{\textbf{\textit{Index terms---}} #1}
\begin{document}
\maketitle

\begin{abstract}
This paper delineates the visual speech recognition (VSR) system introduced by the NPU-ASLP (Team 237) in the second \textbf{C}hi\textbf{n}ese Continuous \textbf{V}isual \textbf{S}peech \textbf{R}ecognition \textbf{C}hallenge (CNVSRC 2024), engaging in all four tracks, including the fixed and open tracks of Single-Speaker VSR Task and Multi-Speaker VSR Task.
In terms of data processing, we leverage the lip motion extractor from the baseline\footnote{https://gitlab.com/csltstu/sunine/-/tree/cncvs} to produce multi-scale video data.
Besides, various augmentation techniques are applied during training, encompassing speed perturbation, random rotation, horizontal flipping, and color transformation.
The VSR model adopts an end-to-end architecture with joint CTC/attention loss, introducing Enhanced ResNet3D visual frontend, E-Branchformer encoder, and Bi-directional Transformer decoder. 
Our approach yields a 30.47\% CER for the Single-Speaker Task and 34.30\% CER for the Multi-Speaker Task, securing second place in the open track of the Single-Speaker Task and first place in the other three tracks.
\end{abstract}

\keywords{Visual Speech Recognition, Lip Reading}

\section{Introduction}


To advance research in visual speech recognition (VSR), the first \textbf{C}hi\textbf{n}ese Continuous \textbf{V}isual \textbf{S}peech \textbf{R}ecognition \textbf{C}hallenge\footnote{http://cnceleb.org/competition} (CNVSRC 2023) was initiated, aiming to probe the performance of large vocabulary continuous visual speech recognition (LVCVSR) in two scenarios: reading in a recording studio and speech on the Internet. 
Furthermore, CNVSRC 2023 employs the CN-CVS~\cite{chen2023cn} dataset as its training set and defines two tasks: Single-Speaker VSR Task (T1) and Multi-Speaker VSR Task (T2). 
Independent development and evaluation sets accompany each task, denoted as CNVSRC-Single.Dev/Eval (T1.Dev/Eval) and CNVSRC-Multi.Dev/Eval (T2.Dev/Eval), respectively. 
Building on the success of CNVSRC 2023, CNVSRC 2024 further delves into Chinese continuous lip reading by offering a more advanced baseline system and releasing a new large-volume Chinese audio-visual speech dataset, named CN-CVS2-P1, for open tracks. 
Additionally, the track settings and datasets, including the final evaluation sets for both tasks, remain consistent with CNVSRC 2023, allowing teams to compare their systems with those in the last session.

This paper describes our system in CNVSRC 2024.
We have continued to use some methods from our best-performing system~\cite{wang2024npu} in CNVSRC 2023, including the data processing workflow, triple speed perturbation, the dynamic data augmentation strategy during training, and the joint CTC/attention model architecture. 
The improvements are as follows. 
First, we expand the maximum video crop size from 112 to 128. 
Second, in the model structure, we upgrade the visual front-end module for extracting video data features by introducing our recently proposed Enhanced ResNet3D~\cite{wang2024enhancing}, which has proven its effectiveness in the ICME 2024 ChatCLR challenge. 
For the decoder part, we adopt a bi-directional Transformer instead of a standard one.
Additionally, we utilize a two-stage training strategy that includes training on the training set and fine-tuning on the development set. 
Finally, using Recognizer Output Voting Error Reduction (ROVER) for post-fusion, we achieve character error rates (CERs) of 30.47\% on T1.Eval and 34.30\% on T2.Eval.
\begin{table*}[ht]
	\centering
	\caption{The CER(\%) results of our VSR systems on Dev and Eval sets in T1 and T2 tasks. Crop refers to the size of training lip video data. Suffix FT represents the decoding result after fine-tuning.}
    \resizebox{\linewidth}{!}{
        \begin{tabular}{c|cccc|ccc|ccc}
    		\toprule
    		System & Visual Frontend & Encoder & Decoder & Crop & T1.Dev & T1.Eval & T1.Eval.FT & T2.Dev & T2.Eval & T2.Eval.FT \\
    		\hline
    		M1 & ResNet3D & E-Branchformer & Transformer & 80 & 41.76 & - & - & 48.50 & - & - \\
    		M2 & ResNet3D & Conformer & Transformer & 96 & 39.27 & - & - & 46.21 & - & - \\
            M3 & ResNet3D & Branchformer & Transformer & 96 & 38.86 & - & - & 46.49 & - & - \\
    		M4 & ResNet3D & E-Branchformer & Transformer & 96 & 38.39 & - & - & 45.97 & - & - \\
            M5 & Enhanced ResNet3D & E-Branchformer & Transformer & 96 & 38.30 & - & - & 45.79 & 45.43 & 41.75 \\
            M6 & ResNet3D & E-Branchformer & Transformer & 112 & 38.13 & - & 35.91 & 44.88 & 44.56 & 40.37 \\
            M7 & Enhanced ResNet3D & E-Branchformer & Bi-Transformer & 112 & 36.60 & 36.79 & 34.56 & 43.29 & 43.13 & 38.73 \\        
            M8 & Enhanced ResNet3D & E-Branchformer & Bi-Transformer & 128 & 36.48 & 37.04 & \textbf{34.27} & 43.23 & 42.99 & \textbf{37.93} \\
            \midrule
            R1 & \multicolumn{4}{c|}{ROVER \textbf{Eval} of M1$\sim$M8} & \multicolumn{3}{c|}{32.7199} & \multicolumn{3}{c}{38.3602} \\
            R2 & \multicolumn{4}{c|}{ROVER \textbf{R1} and \textbf{Eval.FT} of M1$\sim$M8} & \multicolumn{3}{c|}{\textbf{30.4679}} & \multicolumn{3}{c}{\textbf{34.2955}} \\
    		\bottomrule
    	\end{tabular}
    }
	\label{table-1}
\end{table*}
\section{Proposed system}
\begin{figure}[t]
  \centering
  \centerline{\includegraphics[width=8.0cm]{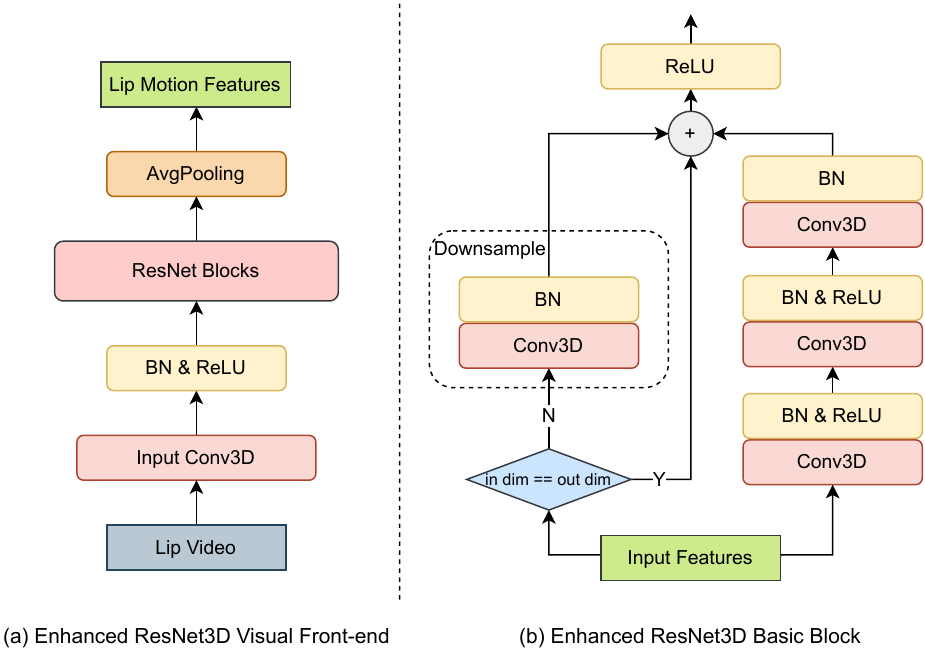}}
\caption{Detailed structures of (a) the Enhanced ResNet3D visual front-end and (b) its basic block.}
\label{fig:1}
\end{figure}
\subsection{Enhanced ResNet3D Visual Front-end}
Figure~\ref{fig:1} shows the detailed structures of our Enhanced ResNet3D visual front-end, which mainly consists of three parts (as shown in Figure 1a), the input Conv3D for mapping the feature channels of the input video data to a higher dimension, AvgPooling for averaging the video height and width dimensions at the final, and the ResNet3D blocks in the middle part modeling the visual features. 
Each layer of the ResNet3D block comprises several basic blocks, as detailed in Figure 1b. 
These blocks are primarily comprised of stacked Conv3D and batch norm, forming the video feature modeling unit. 
Additionally, the first basic block in each ResNet3D block performs a two-fold down-sampling of the input visual features in height and width using convolution instead of Maxpooling operation, and maps to higher feature channels. 
\begin{figure}[t]
  \centering
  \centerline{\includegraphics[width=8.0cm]{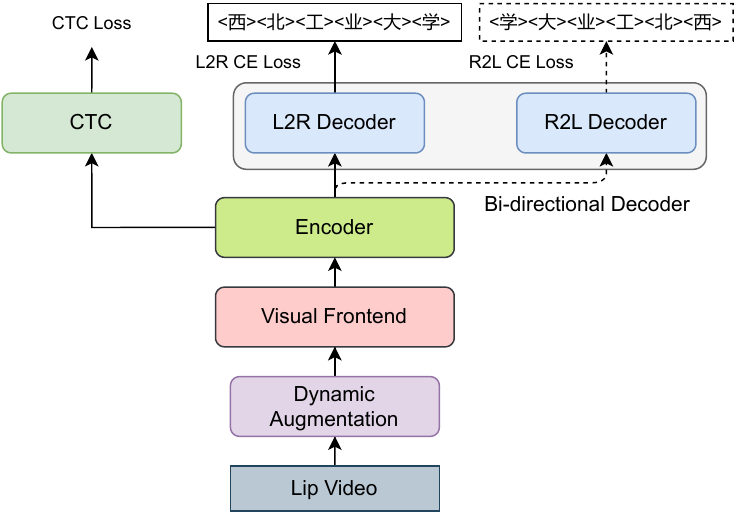}}
\caption{System overview of our VSR system.}
\label{fig:2}
\end{figure}
\subsection{Visual Speech Recognition}
Figure~\ref{fig:2} shows the overall architecture of our VSR system, comprising four main components: dynamic augmentation, visual frontend, encoder, and decoder. 
The training video data is first augmented by random rotation, horizontal flipping, and color transformation through the dynamic augmentation module. 
Subsequently, it passes through the Enhanced ResNet3D visual frontend module for visual feature extraction.
As for the encoder of the VSR system, we build diverse VSR systems featuring different encoders such as Conformer, Branchformer, and E-Branchformer, facilitating multi-system fusion. 
In the decoder part, we use a bi-directional Transformer, which includes a left-to-right (L2R) decoder and a right-to-left (R2L) decoder, as shown in Figure~\ref{fig:2}. 
The final model loss is composed of CTC loss derived from the encoder, L2R CE loss computed on the L2R decoder, and R2L CE loss computed on the R2L decoder, which can be defined as follows:
\begin{equation}
    loss= \lambda loss_{ctc} + (1-\lambda)(\alpha loss^{r2l}_{ce} + (1-\alpha)loss^{l2r}_{ce}),
\end{equation}
where $\lambda$ and $\alpha$ are hyperparameters, both set to 0.3 in this paper.
\section{Experiment}

\subsection{Setup}
All systems are implemented with the WeNet~\cite{yao2021wenet} toolkit, only using the CN-CVS and the development set of CNVSRC-Single or CNVSRC-Multi.
For the encoder of the VSR system, we use a 12-layer E-Branchformer block, each with 256 attention units, 4 attention heads, and 1024 feed-forward units. 
Additionally, the L2R decoder and R2L decoder both contain 3 Transformer layers, each with 4 attention heads and 2048 feed-forward units. 
All systems use the Enhanced ResNet3D visual front-end for feature extraction.
Four ResNet3D blocks have 3, 4, 6, and 3 basic blocks, respectively, with the feature dimensions of 32, 64, 128, and 256.
To achieve better results by post-fusion, we also build Conformer and Branchformer VSR systems with similar param sizes to the E-Branchformer ones.

\subsection{Training and Inference Procedure}
All systems of both tasks are first trained for 50 epochs using the CN-CVS combined with the respective CNVSRC dataset.
Subsequently, we average 15 models with the lowest dev loss.
Next, starting from the averaged model initialization, we perform 10 epochs of fine-tuning using only the CNVSRC dataset of each task and average all 10 epoch models for inference on the final evaluation set.
During inference, we use the attention-rescoring decoding strategy with 64 beam-size and 0.3 CTC weight.

\subsection{Results}
Table \ref{table-1} shows the outcomes of our VSR systems (M1$\sim$M8) and the system fusion results (R1\&R2) obtained by the ROVER. 
Specifically, in terms of the encoder, the performance of the E-Branchformer surpasses that of the Branchformer and Conformer (M2$\sim$M4). 
Moreover, as the crop size increases from 80 to 128, the performance of our VSR system improves (M1$\sim$M8).
Fine-tuning the system with the task-specific CNVSRC dataset significantly improves performance. 
In Task 1, fine-tuning reduces the CER by approximately 2.5\%, while in Task 2 (T2), it can lead to a CER reduction of up to 5\%. 
Finally, after a two-stage multi-system fusion using ROVER, we achieve CERs of 30.4679\% for Task 1 and 34.2955\% for Task 2 on the evaluation set, ranking second place in the open track of the Single-Speaker Task and first place in the other three tracks.


%

%
\footnotesize
\bibliographystyle{IEEEbib}
\bibliography{BibEntries}

\begin{thebibliography}{1}

\bibitem{chen2023cn}
Chen Chen, Dong Wang, and Thomas~Fang Zheng,
\newblock ``Cn-cvs: A mandarin audio-visual dataset for large vocabulary continuous visual to speech synthesis,''
\newblock in {\em Proc. ICASSP}. IEEE, 2023, pp. 1--5.

\bibitem{wang2024npu}
He~Wang, Pengcheng Guo, Wei Chen, Pan Zhou, and Lei Xie,
\newblock ``The npu-aslp-liauto system description for visual speech recognition in cnvsrc 2023,''
\newblock {\em arXiv preprint arXiv:2401.06788}, 2024.

\bibitem{wang2024enhancing}
He~Wang, Pengcheng Guo, Xucheng Wan, Huan Zhou, and Lei Xie,
\newblock ``Enhancing lip reading with multi-scale video and multi-encoder,''
\newblock {\em arXiv preprint arXiv:2404.05466}, 2024.

\bibitem{yao2021wenet}
Zhuoyuan Yao, Di~Wu, Xiong Wang, Binbin Zhang, Fan Yu, Chao Yang, Zhendong Peng, et~al.,
\newblock ``Wenet: Production oriented streaming and non-streaming end-to-end speech recognition toolkit,''
\newblock {\em arXiv preprint arXiv:2102.01547}, 2021.

\end{thebibliography}

\end{document}